\def\year{2015}
\documentclass[letterpaper]{article}
\usepackage{aaai}
\usepackage{times}
\usepackage{helvet}
\usepackage{courier}
\usepackage{float}
\usepackage{amsmath}
\usepackage{subcaption}
\usepackage{pdfpages}

\usepackage[linesnumbered,ruled,vlined]{algorithm2e}

\newcommand{\bcubed}{B$^3$RTDP }

\newcommand{\argmin}{\operatornamewithlimits{argmin}}
\newcommand{\argmax}{\operatornamewithlimits{argmax}}
\newcommand \sigginame {Sigurdur Orn Adalgeirsson}

\nocopyright

\begin{document}

\title{B$^3$RTDP: A Belief Branch and Bound Real-Time\\Dynamic Programming Approach to Solving POMDPs}
\author{\sigginame\\
MIT Media Lab\\
\And
Cynthia Breazeal\\
MIT Media Lab\\
}

\maketitle
\begin{abstract}
\begin{quote}
Partially Observable Markov Decision Processes (POMDPs) offer a promising world representation for autonomous agents, as they can model both transitional and perceptual uncertainties. Calculating the optimal solution to POMDP problems can be computationally expensive as they require reasoning over the (possibly infinite) space of beliefs. Several approaches have been proposed to overcome this difficulty, such as discretizing the belief space, point-based belief sampling, and Monte Carlo tree search. The Real-Time Dynamic Programming approach of the RTDP-Bel algorithm approximates the value function by storing it in a hashtable with discretized belief keys. We propose an extension to the RTDP-Bel algorithm which we call Belief Branch and Bound RTDP (B$^3$RTDP). Our algorithm uses a bounded value function representation and takes advantage of this in two novel ways: a search-bounding technique based on action selection convergence probabilities, and a method for leveraging early action convergence called the \textit{Convergence Frontier}. Lastly, we empirically demonstrate that B$^3$RTDP can achieve greater returns in less time than the state-of-the-art SARSOP solver on known POMDP problems.
\end{quote}
\end{abstract}

\section{Introduction}
As autonomous agents and robots face increasingly challenging planning and reasoning tasks, it is important that they can adequately represent and reason about their capabilities and surroundings. Markov Decision Processes (MDPs) \cite{bellman1957markovian} have been heavily used as a core representation for probabilistic planning problems and have proven sufficiently powerful for many domains, but lack expressivity to represent uncertainty about one's state or perception. Partially Observable MDPs (POMDPs) \cite{Kaelbling199899} allow for state uncertainty by reasoning about beliefs (probability distributions over states), at their core and using the concept of observations and observation functions.
	
The major problem for the use of POMDPs on practical real-world planning domains has been that because of the expressivity of the model, finding an optimal solution can be very computationally expensive. This is in most part due to the ``curse of dimensionality'' as POMDP algorithms must reason about the (possibly infinite) space of beliefs. Several approaches have been proposed for making this problem more tractable, such as point-based belief sampling \cite{pineau2003point} \cite{kurniawati2008sarsop}, Monte-Carlo tree search \cite{silver2010monte}, and belief discretization \cite{Geffner98solvinglarge} \cite{bonet2009solving}.

We had been experimenting with Real-Time Dynamic Programming (RTDP) algorithms for solving MDP problems and believed there was great potential in applying a similar technique to solving POMDPs. An algorithm RTDP-Bel \cite{Geffner98solvinglarge} had been proposed early for applying the RTDP technique to POMDPs, but hadn't received much attention by the community, despite a more recent demonstration by the original authors \cite{bonet2009solving} showing that it was competitive with the state-of-the-art at the time. We believed that based on this there was great potential for extending and improving the RTDP-Bel algorithms with similar techniques that had been used to improve the convergence time of the original RTDP system. 

In this paper our extention to RTDP-Bel which we call Beleif Branch and Bound RTDP (B$^3$RTDP). This algorithm employs a bounded value function representation and emphasizes exploration towards areas of higher value uncertainty to speed up convergence. We also introduce a novel method of pruning action selection by calculating the probability action convergence and pruning when that probability exceeds a threshold. Lastly we experiment with a novel concept of a \textit{Convergence Frontier} (CF) which takes advantage of action convergence near the root of the search tree which effectively shortens search trials.

We demonstrate empirically on known POMDP benchmarking problems \textit{Rocksample} \cite{smith2004heuristic} and \textit{Tag} \cite{pineau2003point} that \bcubed can acquire higher Aver Discounted Reward (ADR) and converge faster than the state-of-the-art SARSOP planner \cite{kurniawati2008sarsop} and furthermore that its parameters allow the user flexibility trading solution quality with convergence time.

\section{Background}

Since the proposed \bcubed algorithm is based on the RTDP algorithm, and incorporates ideas that were either adapted from, or inspired by, advancements that were made to the original RTDP algorithm for MDPs, we'll start by introducing some of those systems. 

\subsection{RTDP}\label{section:RTDP}

Real-Time Dynamic Programming (RTDP) \cite{barto1995learning} is an early DP algorithm for solving MDPs that combines heuristic search with  asynchronous value function updates. This approach provided significant benefit over existing DP algorithms by performing value updates only on the portion of the state space that is relevant to solving the problem (assuming a good initial heuristic). RTDP is an anytime algorithm that generally produces good policies fast, but its convergence time can be slow. Several extension to the original RTDP algorithm have been proposed to improve its convergence time, all of which attempt to better focus the exploration and value updates towards the most ``fruitful'' parts of the state space.

Labeled-RTDP \cite{bonet2003labeled} (LRTDP) introduced a method of marking states as \textit{solved} once their values and those of all states reachable from them had converged. Solved states would subsequently be avoided in the RTDP's exploration of the state space. 

Bounded-RTDP (BRTDP) \cite{mcmahan2005bounded} introduced using a bounded value function, with both an upper and lower value attached to each state. BRTDP takes advantage of this in two ways, for search iteration termination and search exploration guidance. The BRTDP exploration strategy is motivated to explore areas of large value gaps or uncertainty in the value function to quickly ``collapse'' its boundaries onto the optimal value $V^*$.

Several other algorithms have been proposed to improve RTDP using a bounded value function, each of which providing different search exploration heuristics and iteration termination criteria \cite{sanner2009bayesian} \cite{smith2006focused}.

\subsection{POMDPs}
The RTDP algorithm solves MDP problems but they are limited to only modeling uncertainty in the transition function and assume that the agent knows its state with certainty at all times. This is clearly not the case with agents in the real world and therefore we turn to more expressive model, namely the POMDP. 

A fully specified discrete POMDP is defined by: a finite set of states and actions $\mathcal{S}$ and $\mathcal{A}$, a transition function that defines the transition probabilities $\mathcal{T}(s,a,s')$, a set of observations $\mathcal{O}$, observation function $\Omega(a,s',o)$, a reward function $\mathcal{R}(s,a)$, and a discount factor $\gamma$. The transition function $\mathcal{T}$ encodes the dynamics of the environment. It specifies how each action will (possibly stochastically) transition the current state to the next one. As an example, for a wheeled mobile robot this might model the uncertainty about how its wheels might slip on different surfaces. The observation function models the uncertainty in the agent's perception, for the wheeled mobile robot this could be used to model the uncertainty about it's camera-based localization. The reward function $\mathcal{R}$ can be used to encode the goal of the planning task or more generally to specify states and/or actions that are desirable.

In a partially observable domain, the agent cannot directly observe its state and therefore needs to simultaneously do state estimation with planning. A planning agent represents its uncertainty about its current state as a probability distribution over possible states which we will refer to as a \emph{belief} $b$ where $b(s) := Pr(s|b)$. Given the transition and observation functions, a Bayesian belief update can be calculated where the agent determines what next belief $b_a^o$ to hold if it originally holds belief $b$, takes action $a$, and receives observation $o$. 
\begin{equation}\label{eqn:beliefUpdate}
b_a^o(s') = \frac{\Omega(a,s',o)\sum_{s\in S}\mathcal{T}(s,a,s')b(s)}{Pr(o| b,a)}
\end{equation}
\begin{equation}\label{eqn:beliefUpdateNormalization}
Pr(o|b,a) = \sum_{s'\in S}\Omega(a,s',o)\sum_{s\in S}\mathcal{T}(s,a,s')b(s)
\end{equation}

Using equations \ref{eqn:beliefUpdate} and \ref{eqn:beliefUpdateNormalization}, we can restate the Bellman DP value update equations for POMDPs as such:

\begin{equation}\label{eqn:planning:bellmanPOMDP}
Q(b,a) := \sum_{s\in S}b(s)\mathcal{R}(s,a) + \gamma\sum\limits_{o\in O}Pr(o| b,a)V(b_a^o)
\end{equation}

\begin{equation}\label{eqn:planning:bellmanPOMDP}
V(b) := \max_{a\in A}Q(b,a)
\end{equation}

These equations recursively define the value of a belief $V(b)$ and the value of taking an action in a belief $Q(b,a)$. The latter is defined as the expected reward of taking that action plus the discounted successive belief value expectation, and the former as the maximum $Q(b,a)$ for all actions $a$. The optimal action policy can therefore be defined as always choosing the action with the highest $Q$ value, or $\pi^*(b) := \argmax_{a\in A}Q(b,a)$.

\subsection{RTDP-Bel}
Geffner and Bonet presented an extension to the RTDP called RTDP-Bel which is able to handle partially observable domains \cite{Geffner98solvinglarge}. The most significant difference between RTDP and RTDP-Bel (in addition to the fact that the former searches over state-action graphs and the latter over belief-action-observation graphs) is the way in which the value function is represented.

Value functions over beliefs are much more challenging to implement than for discrete states as for any finite set of states there is an infinite set of beliefs. One of the most commonly used representations (attributed to Sondik \cite{sondik1971optimal}) maintains a set of $\alpha$ vectors, each of dimension $|S|$, where $V(b)=\max_{\alpha}\alpha\cdot b$. RTDP-Bel uses a function-approximation scheme which discretizes the beliefs and stores their values in a hashtable which uses the discretized belief as key.

\begin{equation}
\hat{b}(s) = ceil(D\cdot b(s))
\end{equation}

The discretization parameter $D$ therefore determines how ``close'' two beliefs should be in their assignment of probabilities to states, so that their values should be tconsidered the same. The value function for RTDP-Bel is therefore defined as follows:

\begin{equation}\label{eqn:planning:discretizedValueFunction}
\hat{V}(b)=\left\{
\begin{array}{l l l}
	h(b) & | & \text{if}\; \hat{b} \notin \textsc{HashTable}\\
	\textsc{HashTable}(\hat{b}) & | & otherwise \\
\end{array}\right.
\end{equation}

Where $h(b)$ is defined as $\sum_{s\in b}{b(s)h(s)}$ and $h$ is an admissible heuristic for this problem.

\begin{algorithm}[h!]
	\SetKwFunction{RTDPBel}{RTDP-Bel}
	\SetKwProg{RTDPBel}{\textsc{RTDP-Bel}}{}{}
	\RTDPBel{$(b_I: Belief)$}{
	\While{\textbf{not} \textsc{terminate}}{
		\nl $b := b_I; \ \ s \sim b$\;
		\While{$b \neq \textsc{goal}$}{
			\nl $a := \argmin_{a'\in A_{b}}\hat{Q}(b,a')$\;
			\nl $\hat{V}(b) := \hat{Q}(b,a)$\;
			\nl $s' \sim \mathcal{T}(s,a,\cdot)$\;
			\nl $o \sim \Omega(a,s',\cdot)$\;
			\nl $b := b_a^o; \ \ s := s'$\;
		}
	}
	}{}

\caption{The RTDP-Bel algorithm}
\label{alg:planning:rtdpbel}
\end{algorithm}

One critical issue with RTDP-Bel is that it offers no reliable termination criteria, one can terminate the algorithm at any time and extract an action policy but without any guarantees that it has converged. 

\subsubsection{Transforming Goal POMDPs to General POMDPs}

One limitation of the RTDP algorithm is that it is only applicable to so-called \textit{Stochastic Shortest Path} MDP problems. Similarly, RTDP-Bel is only applicable to \textit{Goal POMDPs} which satisfy the following criteria: All actions rewards need to be strictly negative and a set of fully observable goal states exist that are absorbing and terminal.
 
These constraints seem on first sight quite restrictive and would threaten to limit RTDP-Bel to only be applicable to a small subset of all possible POMDP problems. But this is not the case as Bonet et al. demonstrate in \cite{bonet2009solving} that any general discounted POMDP can be transformed into a goal POMDP.

\subsection{Other POMDP Planning Approaches}

Algorithms exist to solve for the optimal value function of POMDPs \cite{sondik1971optimal}, but this is rarely a good idea as the belief space is infinitely large and only a small subset of it relevant to the planning problem. A discovery by Sondik about the value function's piece-wise linear and convex properties led to a popular value function representation which consists of maintaining a set of $|S|$-dimensional $\alpha$ vectors, each representing a hyperplane over the state space. This representation is named after its author Sondik and many algorithms, both exact and approximate, take advantage of it.

The \textit{Heuristic Search Value Iteration} (HSVI) algorithm \cite{smith2004heuristic} extends ideas of employing heuristic search from \cite{Geffner98solvinglarge} and combines them with the Sondik value function representation but with upper and lower bound estimates. It employs information seeking observation sampling technique akin to that of BRTDP (introduced below) but aimed at minimizing \textit{excess uncertainty}. The \textit{Point-based Value Iteration} (PBVI) algorithm \cite{pineau2003point} doesn't use a bounded value function but introduced a novel concept of maintaining a finite set of relevant belief-points and only perform value updates for those sampled beliefs.

Lastly the SARSOP algorithm \cite{kurniawati2008sarsop} combines the techniques of HSVI and PBVI, performing updates to its bounded Sondik-style value function over a finite set of sampled belief points. Additionally, SARSOP uses a novel observation sampling method which uses a simple learning technique to predict which beliefs should have higher sampling probabilities to close the value gap on the initial belief faster.

\section{Belief Branch and Bound RTDP}

In this section we present a novel planning algorithm for POMDPS called \textit{Belief Branch and Bound Real-Time Dynamic Programming} (\bcubed). The algorithm extends the RTDP-Bel system with a bounded value function, a \textit{Branch and Bound} style search tree pruning and influences from existing extensions to the original RTDP algorithm for MDPs.

\subsection{Bounded Belief Value Function}\label{section:planning:valueFunction}

As was previously mentioned, \bcubed maintains a bounded value function over the beliefs. In the following discussion, I will refer to these boundaries as separate value functions $\hat{V}_L(b)$ and $\hat{V}_H(b)$ but the implementation actually stores a two-dimensional vector of values for each discretized belief in the hash-table (see equation \ref{eqn:planning:discretizedValueFunction}) and so only requires a single lookup operation for the value retrieval of both boundaries.

It is desirable to initialize the lower bound of the value function to an admissible heuristic for the planning problem. This requirement needs to hold for an optimality guarantee. It is easy to convince oneself of why this is, imagine that at belief $b_1$, all successor beliefs to taking the optimal action $a_1$ have been improperly assigned inadmissible heuristic values (that are too high). This will result in an artificially high $Q(b_1,a_1)$ value, resulting in the search algorithm choosing to take $a_2$. If we assume that the successor beliefs of action $a_2$ were initialized to an admissible heuristic then after some number of iterations we can expect to have learned the true value of $Q^*(b_1,a_2)$, but if that value is still lower than the (incorrect) $Q(b_1,a_1)$, then we will never actually choose to explore $a_1$ and never learn that it is in fact the optimal action to take.

It is equally desirable to initialize the upper boundary $\hat{V}_H(b)$ to a value that overestimates the cost of getting to the goal. This becomes evident when we start talking about the \textit{bounding} nature of the \bcubed algorithm, namely that it will prune actions whose values are dominated with a certain confidence threshold. For that calculation, we require that the upper boundary be an admissible upper heuristic for the problem.

In the previous section, we discussed a method of how to transform \textit{discounted POMDPs} to \textit{goal POMDPs}. This transformation is particularly useful when working with \textit{goal POMDPs} we get theoretical boundaries on the value function for free. Namely, there is no action that has zero or negative costs which means that no belief, other than goal beliefs, has zero or negative values. This means that the heuristic $h_L(b) = 0$ is an admissible (although not very informative) heuristic. Similarly we know that since the domain is effectively discounted (through the artificial transition with probability $1-\gamma$ to the artificial goal state) that the absolute worst action policy an agent could follow would be repeatedly taking the action with highest cost. Because of the discounted nature of the domain, this (bad) policy has a finite expected value of $\max(\mathcal{C}(s,a))/(1-\gamma)$ which provides a theoretical upper bound. This value bound is called the \textit{Blind Action} value and was introduced by \cite{Hauskrecht:2000:VAP:1622262.1622264}.

Even though it is important that the heuristics (both lower and upper) be admissible, and it is nice that we can have guaranteed ``blind'' admissible values, it is still desirable that the heuristics be informative and provide tighter value bounds. Uninformed heuristics can require exhaustive exploration to learn which parts of the belief space are ``fruitful''. An informed heuristic can quickly guide a search algorithm towards a solution that can be incrementally improved. This is especially important for RTDP-based algorithms as they effectively search for regions of good values and then propagate those values back to the initial search node through \textit{Bellman} updates. What this effectively means is that the sooner the algorithm finds the ``good'' part of the belief space, the quicker it will converge.

Since the lower boundary of the value function is used for exploration, the informative quality of the lower heuristic plays a much bigger role in the convergence time of the algorithm.

Domain-dependent heuristics can be hand-coded by domain experts, which is a nice option to have when the user of the system possesses a lot of domain knowledge that could be leveraged to solve the problem. In lieu of good domain-dependent heuristics, we require methods to extract domain-independent heuristics that are guaranteed to be admissible.

There are several different ways to obtain admissible lower heuristics from a problem domain. The most common method is called the $Q_{MDP}$ approach and was introduced by \cite{littman1995learning}. This approach ignores the observation model in the POMDP and simply solves the MDP problem defined by the transition and cost/reward model specified. This problem can be solved with any MDP solver much faster than the full POMDP and provides a heuristic value for each state in the domain which can be combined into a belief heuristic as such: $h(b) = \sum_{s\in S} h(s)b(s)$. The $Q_{MDP}$ heuristic provides an admissible lower bound to the POMDP problem as it is solving a strictly easier fully observable MDP problem. This heuristic tends to work well in many different domains, but it generally fails to provide a good heuristic in information-seeking domains since it completely ignores the observation function.

Another admissible domain-independent heuristic is the \textit{Fast Informed Bound} (FIB) which was developed by Hauskrecht \cite{Hauskrecht:2000:VAP:1622262.1622264}. This heuristic is provably more informative than $Q_{MDP}$ as it incorporates the observation model of the domain. This more informative heuristic does come at a higher cost of $O(|A|^2|S|^2|O|)$ whereas $Q_{MDP}$ has the complexity of regular \textit{Value Iteration} or $O(|A||S|^2)$.

There also exist methods to improve the upper bound of the value function. One method is to use a point-based approximation POMDP solver to approximately solve the actual POMDP problem as long as the approximation is strictly over-estimating \cite{ross2008online}. This is a very costly operation but can be worth it for certain domains.

The \bcubed algorithm can be initialized to use any of the above or more heuristic strategies, but we have empirically found its performance satisfactory when initialized with the $Q_{MDP}$ heuristic as the lower bound and the \textit{blind worst action policy} heuristic as its upper bound.

\subsection{Calculating Action Selection Convergence}

A central component of the \bcubed algorithm is determining when the search tree over beliefs and actions can be pruned. This pruning will lead to faster subsequent \emph{Bellman} value updates over the belief in question, less memory to store the search tree, and quicker greedy policy calculation.

\begin{figure}[h!]
\centering
\begin{subfigure}{1.7in}
  \centering
  \includegraphics[width=1.7in]{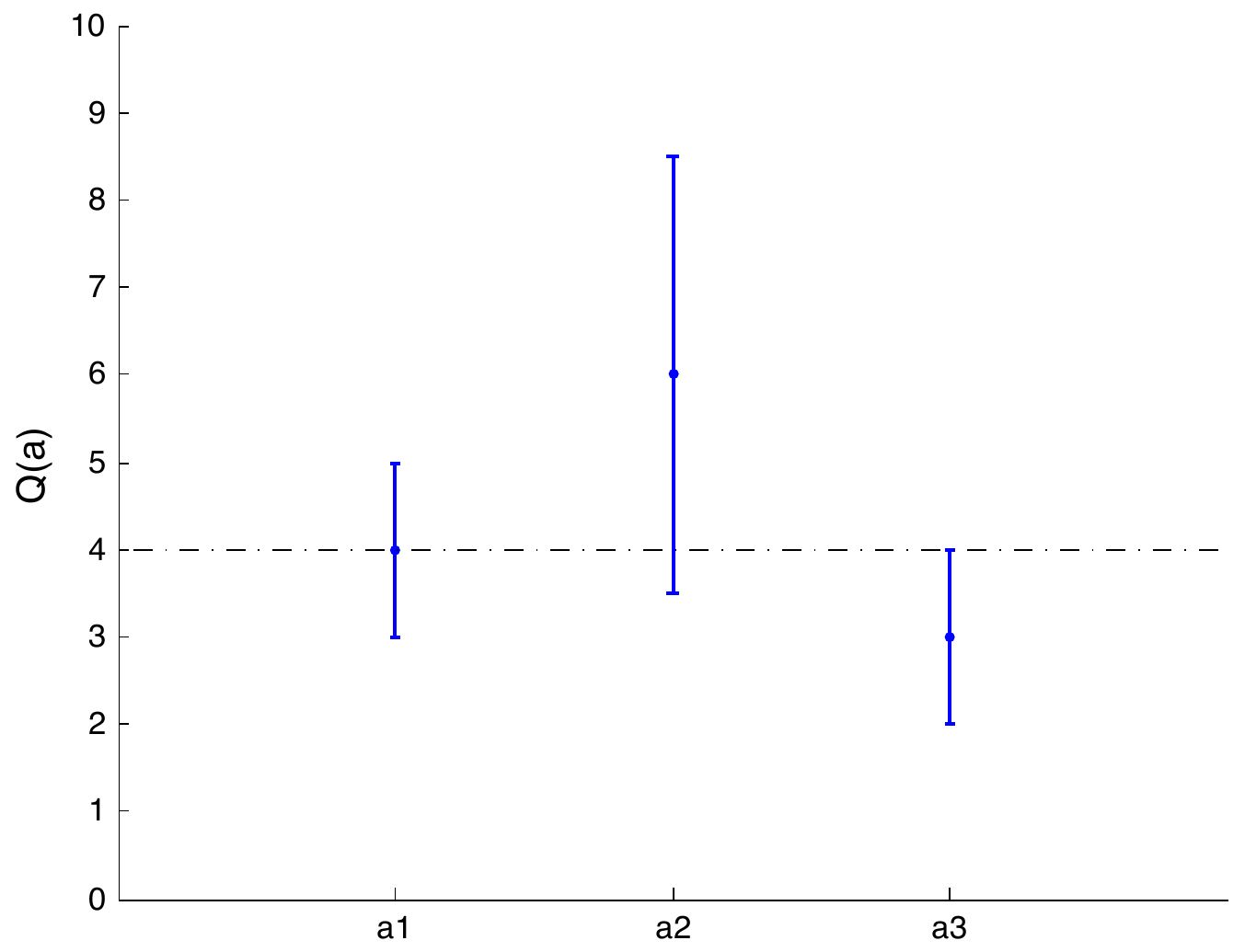}
  \parbox{1.6in}{\caption{Demonstrates example Q boundaries for three actions. Action $a_3$ is the greedy action to choose and it seems like its Q value will dominate that of $a_2$}}
\end{subfigure}%
\begin{subfigure}{1.7in}
  \centering
  \includegraphics[width=1.7in]{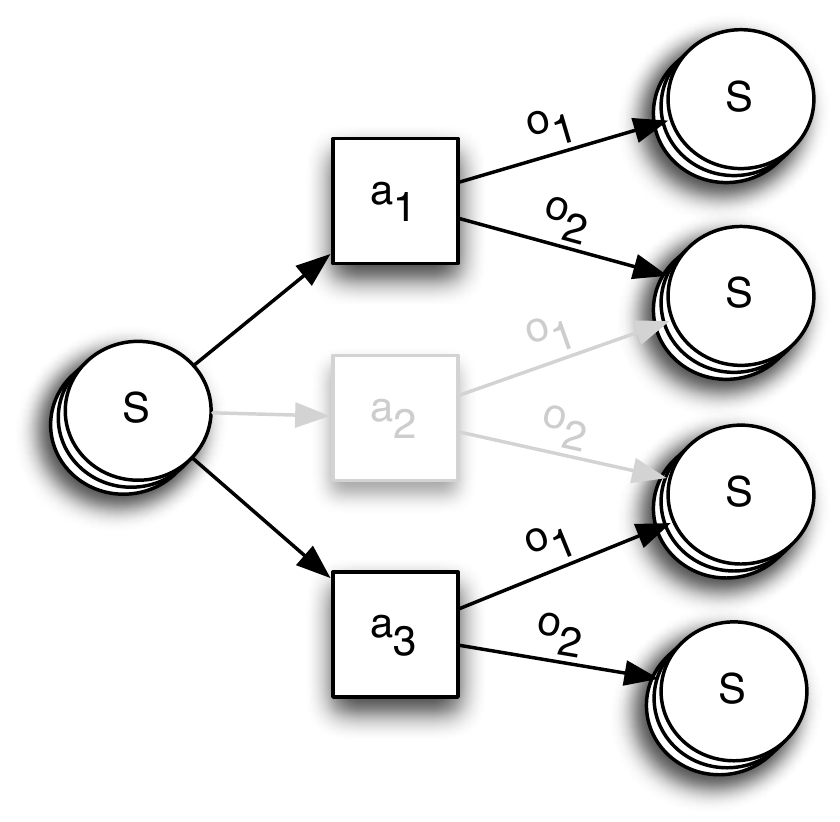}
  \parbox{1.6in}{\caption{Beliefs and actions relevant to the figure on the left}}
\end{subfigure}%
\caption{}
\label{fig:planning:fakeQValues}
\end{figure}

Traditionally, \textit{Branch and Bound} algorithms only prune search nodes when their bounds are provably dominated by the bounds of a different search node (therefore making traversal of the node a sub-optimal choice). To mitigate the large belief space that POMDP models can generate we experiment with pruning actions \textit{before} they are actually provably dominated. Figure \ref{fig:planning:fakeQValues} demonstrates how the search algorithm might find itself in a position where it could be quite certain (within some threshold) that one action dominates another but it could still cost many search iterations to make absolutely sure.

We use the assumption that the true value of a belief is uniformly distributed between the upper and lower bounds of its value. Notice that the following calculations could be carried out for any number of types of value distributions, but the uniform distribution is both easier to calculate in closed form and appropriate since we really do not have evidence to support a choice of a differently shaped distribution.

\begin{figure}[h!]
\centering
\includegraphics[width=2.7in]{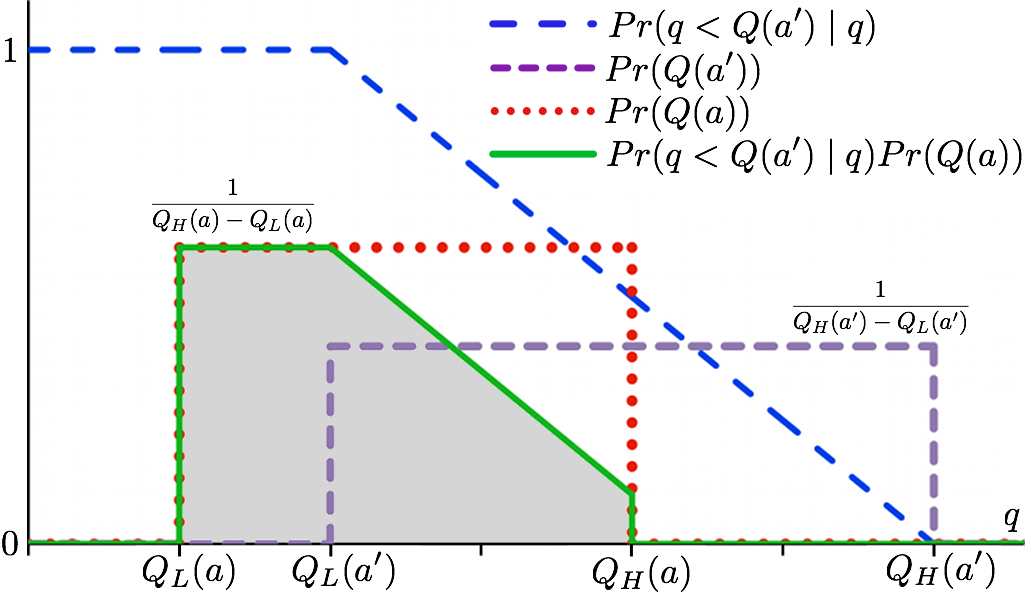}
\caption{Finally this figure shows the function whose integral is our quantity of interest $Pr(Q^*(a)<Q^*(a'))$. This integral will always simply be the sum of rectangle and triangle areas for two uniform Q distributions.}
\label{fig:planning:qcalc4}
\end{figure}

If we assume that the true value of a belief is uniformly distributed between its bounds, then the action Q values are also uniformly distributed and the following holds:

\begin{equation}\label{equation:uniformDist}
Pr(q=Q^*(a))=\left\{
\begin{array}{l l l}
	\frac{1}{G(a)} & | & Q_L(a)<q<Q_H(a)\\
	0 & | & otherwise \\
\end{array}\right.
\end{equation}

\begin{equation}\label{equation:probQLowerThanQstar}
Pr(q<Q^*(a) \,\vert\, q)=\left\{
\begin{array}{lll}
	1 & | & q<Q_L(a)\\
	\frac{Q_H(a)-q}{G(a)} & | & Q_L(a)<q<Q_H(a)\\
	0 & | & q>Q_H(a) \\
\end{array}\right.
\end{equation}

We are interested in knowing the probability that one action's Q value is lower than another's at any given time during the runtime of the algorithm so that we can determine whether or not to discard the latter. This is a crucial operation to the \emph{bounding} portion of the algorithm. The quantity we are interested is therefore $Pr(Q^*(a) < Q^*(a'))$ when calculating whether we can prune action $a'$ since its Q value is dominated by that of $a$.

We start by noticing that there are two special cases that can be quickly determined in which the quantity of interest is either 0 or 1. If $Q_H(a)\leq Q_L(a')$ then all of the probability mass of $Q^*(a)$ is guaranteed to be below that of $Q^*(a')$ and therefore $Pr(Q^*(a) < Q^*(a')) = 1$. By the same rationale we have $Pr(Q^*(a) < Q^*(a')) = 0$ when $Q_H(a')\leq Q_L(a)$.

\begin{equation}\label{equation:convergenceCalculation1}
Pr(Q^*(a) < Q^*(a')))=\left\{
\begin{array}{l l l}
	0 & | & Q_H(a')\leq Q_L(a)\\
	1 & | & Q_H(a)\leq Q_L(a') \\
	\text{eqn. \ref{equation:convergenceCalculation2}} & | & otherwise\\
\end{array}\right.
\end{equation}

We apply equations \ref{equation:uniformDist}, \ref{equation:probQLowerThanQstar}, and the law of total probability to obtain the following:

\begin{multline}\label{equation:convergenceCalculation2}
\int\limits_{Q_L(a)}^{Q_H^{min}} \! Pr(q<Q^*(a') \,\vert\, q)Pr(q=Q^*(a)) \, \mathrm{d}q = \\
\frac{Q_L(a')-Q_L(a)}{Q_H(a)-Q_L(a)} + \\
\frac{2Q_H(a')Q_H^{min}-2Q_H(a')Q_L(a') - \left(Q_H^{min}\right)^2+\left(Q_L(a')\right)^2}{2(Q_H(a)-Q_L(a))(Q_H(a')-Q_L(a'))}
\end{multline}

This calculation assumes that $Q_L(a)<Q_L(a')$ (which is the only case in which we would be interested in performing it) and uses the shorthand notation $Q_H^{min} = \min(Q_H(a),Q_H(a'))$. The outcome can be arrived at in a more graphically intuitive manner which is demonstrated in figure \ref{fig:planning:qcalc4}. It should be clear that the assumption of a uniform distribution of the $Q*$ value between its bounds leads to this rather efficient calculation of simply accumulating areas of rectangles and triangles.

\subsection{Convergence Frontier}\label{section:planning:convergenceFrontier}

\begin{figure}[h!]
\centering
\includegraphics[width=2.7in]{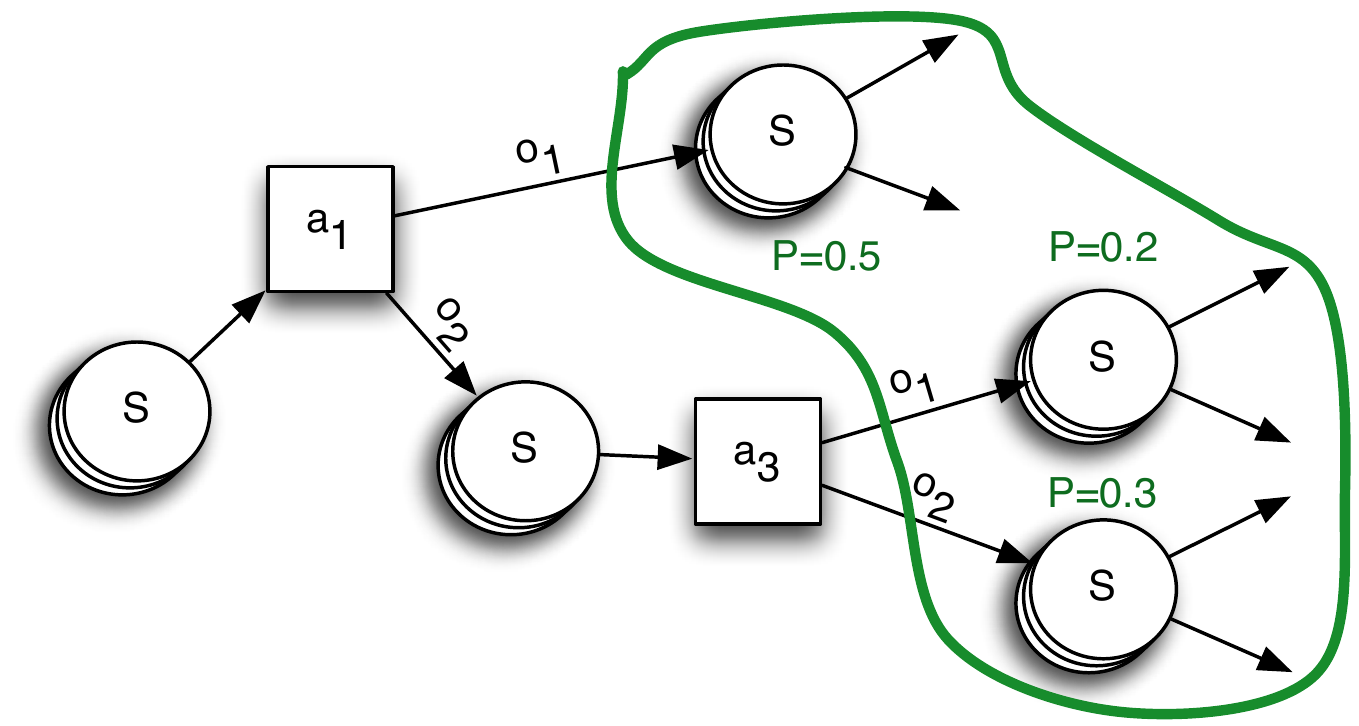}
\caption{Demonstrates how action choice can converge over a belief, creating effectively a frontier of reachable successive beliefs with associated probabilities. This effect can be taken advantage of to shorten planning.}
\label{fig:planning:convergenceFrontier}
\end{figure}

The \textit{Convergence Frontier} (CF) is a concept created to take advantage of early action convergence close to the initial search node. An intuitive understanding of it can be gained by thinking about when the action choice for a belief has converged to only one action that dominates all others. At that point, actually simulating action selection is unnecessary as it will always evaluate to this converged action. This can be the case for several successive beliefs. Figure \ref{fig:planning:convergenceFrontier} demonstrates how action choice can converge over the initial belief as well as some of its successor beliefs, effectively extending the CF further out whenever the action policy over a belief within it converges. When planning for a given POMDP problem, the usefulness of the convergence frontier depends on the domain-dependent difficulty of choosing actions early on.

The CF is initialized to only contain the initial belief with probability one. Whenever the action policy converges to one best action over any belief in the CF, that belief is removed and the successor beliefs of taking that action are added with their respective observation probabilities weighted by the original CF probability of the originating belief. Sometimes the value function converges over a belief before the action policy has converged. In this case, we simply remove this belief from the CF (effectively reducing the total CF probability sum from one) as the value function has been successfully learned at that node. This presents two separate termination criteria: 1. If the total probability of the CF falls below a threshold. And 2. If the total probability weighted value gap of all beliefs in the CF falls below a threshold.

Pseudocode for the \textsc{updateCF} routine is provided in algorithm \ref{alg:planning:bcubed:convergenceFrontier}. The routine iterates through every belief $b$ currently in the frontier, in line 5 it checks to see if the value function has been collapsed over $b$ and if so it is simply removed (which reduces the total probability of the frontier since no subsequent beliefs are added for that node). In line 10 the routine checks to see if only one action is left to be taken for $b$ (the action policy has converged), if so then $b$ is removed from the frontier and all subsequent beliefs of taking the converged action from $b$ are added with their respective observation probabilities multiplied by the probability of $b$ in the frontier.

The short \textsc{shouldTerminateCF} routine is also defined in algorithm \ref{alg:planning:bcubed:convergenceFrontier}. It simply dictates that the algorithm should terminate when either of two conditions are satisfied. The first condition activates when the total probability of all beliefs in the frontier goes below a threshold $\beta$. This means that when acting on the optimal policy, starting at the initial belief, it is sufficiently unlikely that any frontier belief is experienced. The second condition activates when the probability weighted value gap of the frontier goes below the threshold $\epsilon$, this means that the value function has been sufficiently learned over all the beliefs in the frontier.

Lastly, the \textsc{sampleCF} routine in algorithm \ref{alg:planning:bcubed:convergenceFrontier} simply creates a probability distribution by normalizing the probability weighted value gaps of the beliefs in the frontier. At line 33 this distribution is sampled and the corresponding belief returned.

\subsection{Belief Branch and Bound RTDP}

In the previous sections we have introduced many of the relevant important concepts that we now combine together into a novel POMDP planning system called Belief Branch and Bound RTDP (\bcubed). The algorithm extends the RTDPB-Bel \cite{Geffner98solvinglarge} system but uses a bounded value function. It follows a similar belief exploration sampling strategy as the BRTDP \cite{mcmahan2005bounded} system does for MDPs except adapted to operate on beliefs rather than states. This exploration heuristic chooses to expand the next belief node that has the highest promise of reduction in uncertainty. This approach realizes a search algorithm that is motivated to ``seek'' areas where information about the value function can be gained. An interesting side-effect of this search algorithm is that the algorithm never visits beliefs whose values are known ($\hat{V}_L(b)=\hat{V}_H(b)$) such as the goal belief, because there is nothing to be gained by visiting them. Finally \bcubed is a \textit{Branch and Bound} algorithm, meaning that it leverages its upper and lower bounds to determine when certain actions should be pruned out of the search tree.

The \bcubed system is described in detail in algorithms \ref{alg:planning:bcubed} and \ref{alg:planning:bcubed:subroutines}, but we will informally describe its general execution. Initially, the upper bounds on the value function $\hat{V}_H$ are initialized to a \textit{Blind Policy} value \cite{Hauskrecht:2000:VAP:1622262.1622264}, which can generally be easily determined from the problem parameters (namely the discount factor $\gamma$ and the reward/cost function $\mathcal{R}/\mathcal{C}$). The lower bounds of the value function are initialized to an admissible search heuristic such as the $Q_{MDP}$ which can be efficiently calculated by solving the MDP that underlies the POMDP in question by ignoring the observation model.

The initial belief $b_I$ is added to the \textit{Convergence Frontier} (CF discussed in section \ref{section:planning:convergenceFrontier}) with probability one in line 5. Until convergence as determined by the \textsc{shouldTerminateCF} routine in algorithm \ref{alg:planning:bcubed:convergenceFrontier}, the \bcubed algorithm samples an initial belief for that trial $b_T$ from the CF in line 8, performs a \textsc{b$^3$rtdpTrial} and finally updates the CF in line 12.

Each \textsc{b$^3$rtdpTrial} in algorithm \ref{alg:planning:bcubed} initializes a stack of visited beliefs in line 19 and proceeds to execute a loop until either maximum search depth is achieved or the termination criteria (discussed below) is met. On every iteration we push the current belief onto the stack and the find the currently ``best'' action to take according to the lower boundary of the $\hat{Q}_L$ function is found in line 28. To find this action we need to calculate the $\hat{Q}_L$ values for all actions so we might as well use them to perform a \textit{Bellman} value update on the current belief in line 32 (we also perform a bellman update for the upper bound of the value function). We then select the next successor belief to explore, using the \textsc{pickNextBelief} routine line 38. Once the loop terminates, we perform \textit{Bellman} updates on both value boundaries (lines 55 and 57) and action pruning (line 51) on all the beliefs we visited in this trial, in reverse order. This is done to both propagate improved value boundaries back to the initial belief as well as make successive search iterations more efficient.

The \textsc{pickNextBelief} routine in algorithm \ref{alg:planning:bcubed:subroutines} starts by creating a vector containing the observation probability weighted value gaps of the successive beliefs of taking action $a$ in line 3. The sum of the values of this vector is called $G$ and is used both for normalization and determining termination (line 8). If $G$, which is the expected value gap of the next belief to be experienced, is lower than a certain portion (defined by $\tau$) of the value gap at the trial initial belief, then the trial is terminated. Otherwise we sample a value from the normalized vector and return the associated belief.

Lastly, the \textsc{prune} routine in algorithm \ref{alg:planning:bcubed:subroutines} simply iterates through the set of all actions available at the current belief, calculates the probability of them being dominated by the currently best action (line 19), and removes them from the set if that probability is higher than a threshold $\alpha$.

The following are the important parameters for \bcubed along with a discussion of their impact on the algorithm:
\begin{itemize}
	\item $D$: Belief discretization factor. Determines how densely the belief space is clustered for value updates. If set too low then belief clustering might cluster together beliefs that should receive very different values and negatively impact planning result. If set too high then value hash-table will grow very large and many updates will be required to learn the true values of beliefs. Typical range: $[5,20]$. See equation \ref{eqn:planning:discretizedValueFunction}.
	\item $\alpha$: Action convergence probability threshold. This parameter determines when it is appropriate to prune an action from the selection at a given belief. If the $Q^*(b,a_1)$ dominates $Q^*(b,a_2)$ with probability higher than $\alpha$, then $a_1$ is pruned. Typical range: $[0.65, 1]$. See equation \ref{equation:convergenceCalculation}.
	\item $\epsilon$: Minimum value gap. This threshold dictates whether the search algorithm has converged when $\hat{V}_H(b_I)-\hat{V}_L(b_I) < \epsilon$ or when the probability weighted value gap of the \textit{Convergence Frontier} beliefs is below $\epsilon$. Typical range $[0.0001, 0.1]$.
	\item $\beta$: Minimum \textit{Convergence Frontier} probability. This provides a secondary termination criteria for the algorithm. When the total probability of the CF falls below $\beta$ the algorithm terminates. Typical range $[0.0001, 0.01]$.
	\item $\tau$: Trial termination ratio. This parameter is used to determine whether a search trial should be terminated. When the search arrives at a belief where the expected value gap of the successor beliefs is lower than a ratio measured by $\tau$ of the value gap at the trial's initial belief then the iteration is terminated and value updates are propagated backwards to the trial's initial belief. Typical range $[5, 100]$.
\end{itemize}

The parameters that have the biggest impact on the efficiency of \bcubed are $D$ and $\alpha$. For all of the evaluations and future discussion we will use the values $\epsilon = 0.01$, $\beta = 0.001$ and $\tau = 10$, and show results with varying values of $D$ and $\alpha$.

\begin{algorithm}[h!]
	\SetKwProg{updateCF}{\textsc{updateCF}}{}{}
	\updateCF{$(c: CF)$}{
	\ForEach{$b \in c$}{
		\nl \tcp{If value certain, remove it}
		\If{$\hat{V}_H(b)-\hat{V}_L(b)<\epsilon$}{
			\nl $c.\textsc{remove}(b)$\;
		}
		\nl \tcp{If converged, add successors}
		\ElseIf{$\textsc{size}(A(b)) = 1$}{
			\nl $a := \textsc{pickAction}(A(b))$\;
			\ForEach{$o\in O\mid Pr(o|b,a) > 0$}{
				\If{$b_a^o \notin c$}{
					\nl $c.\textsc{append}(b_a^o)$\;
				}
				\nl $c.Pr(b_a^o) += c.Pr(b)\cdot Pr(o|b,a)$\;
			}
			\nl $c.\textsc{remove}(b)$\;
		}
	}
	}{}

	\SetKwProg{terminateCF}{\textsc{terminateCF}}{}{}
	\terminateCF{$(c: CF)$}{
		\Return $\left(\sum\limits_{b\in c} c.Pr(b) < \beta \right) \lor \left(\sum\limits_{b\in c} c.Pr(b)\left(\hat{V}_H(b)-\hat{V}_L(b)\right) < \epsilon \right)$\;
	}{}

	\SetKwProg{sampleCF}{\textsc{sampleCF}}{}{}
	\sampleCF{$(c: CF)$}{
		\nl \tcp{Sample towards high uncertainty}
		\nl $\forall b\in c, g(b) := c.Pr(b)\left(\hat{V}_H(b)-\hat{V}_L(b)\right)$\;
		\nl $G := \sum g$\;
		\nl \KwRet $b'\sim g(\cdot)/G$\;
	}{}
\caption{Convergence Frontier}
\label{alg:planning:bcubed:convergenceFrontier}
\end{algorithm}

\begin{algorithm}[h!]
	\SetKwProg{pickNextBelief}{\textsc{pickNextBelief}}{}{}
	\pickNextBelief{$(b,b_T: Belief, a: Action)$}{
		\nl $\forall o\in O, g(b_a^o) := Pr(o|b,a) (\hat{V}_H(b_a^o)-\hat{V}_L(b_a^o))$\;
		\nl $G := \sum g$\;
		\If{$G < \left(\hat{V}_H(b_T)-\hat{V}_L(b_T)\right)/ \tau$}{
			\nl \textbf{return} $\emptyset$\;
		}
		\nl \KwRet $b'\sim g(\cdot)/G$\;
	}{}
	\SetKwProg{prune}{\textsc{prune}}{}{}
	\prune{$(b: Belief, a_{best}: Action,  A_b: ActionSet)$}{
	\ForEach{$a \in A_b \mid a \neq a_{best}$}{
		\nl \tcp{Obtained by equation \ref{equation:convergenceCalculation1}}
		\If{$Pr\left(Q^*(b,a_{best})) < Q^*(b,a)\right) > \alpha$}{
			\nl $A_b := A_b \setminus a$\;
		}
	}}{}
\caption{Subroutines of the \bcubed algorithm. \textsc{pickNextBelief} defines how the algorithm chooses the next belief to explore. \textsc{prune} shows how action choises can be pruned when near convergence (see figure \ref{fig:planning:qcalc4} for calculation of probability)}
\label{alg:planning:bcubed:subroutines}
\end{algorithm}

\begin{algorithm}[h!]
	\SetKwFunction{BBBRTDP}{BBBRTDP}

	\SetKwProg{BBBRTDP}{\textsc{B$^3$RTDP}}{}{}
	\BBBRTDP{$(b_I: Belief)$}{
	\nl $\textsc{initializeCF}(c, b_I)$\;
	\While{\textbf{not} $ \textsc{terminateCF}(c)$}{
		\nl $b := \textsc{sampleCF}(c)$\;
		\nl $\textsc{b$^3$rtdpTrial}(b)$\;
		\nl $\textsc{updateCF}(c)$\;
	}
	\nl \KwRet $\textsc{GreedyUpperValuePolicy}()$\;}{}

	\SetKwProg{bbbrtdpTrial}{\textsc{b$^3$rtdpTrial}}{}{}
	\bbbrtdpTrial{$(b_T: Belief)$}{
	\nl $visited := \{\}$\;
	\nl $b := b_T$\;
	\While{$(\textsc{size}(visited) \le MAX_{depth}) \land (b \neq \emptyset)$}{
		\nl $\textsc{push}(visited,b)$\;
		\nl \tcp{Pick action greedily}
		\nl $a := \argmax_{a'\in A_{b}}\hat{Q}_H(b,a')$\;
		\nl $b := \textsc{pickNextBelief}(b,b_T,a)$\;
	}
	\While{$\textsc{size}(visited) > 0$}{
		\nl $b := \textsc{pop}(visited)$\;
		\nl $a := \argmax_{a'\in A_{b}}\hat{Q}_H(b,a')$\;
		\nl $\textsc{prune}(b,a,A_{b})$\;
		\nl \tcp{Bellman value update}
		\nl $\hat{V}_H(b) := \hat{Q}_H(b,a)$\;
		\nl $\hat{V}_L(b) := \max_{a'\in A_b}\hat{Q}_L(b,a')$\;
	}}{}
\caption{The Belief Branch and Bound RTDP (\bcubed) algorithm}
\label{alg:planning:bcubed}
\end{algorithm}

\section{Empirical Results}

In this section we present evaluation results for the \bcubed POMDP planning algorithms on two well-known evaluation domains. We have chosen to include evaluation results for a state-of-the art POMDP planner called SARSOP \cite{kurniawati2008sarsop} which is a popular belief point-based heuristic search algorithm. The following points are good to keep in mind when comparing performances between the SARSOP planner and \bcubed:
\begin{enumerate}
  \item Because of the RTDP-Bel \textit{hashtable} value function implementation, \bcubed consumes significantly more memory than SARSOP.
  \item SARSOP takes advantage of the factored structure in the problem domains. This makes for a significantly more efficient belief update calculation. There is no reason why \bcubed could not do the same but it simply hasn't been implemented yet (see section \ref{section:planning:futureWork}).
  \item Even though both algorithms are evaluated on the same machine, SARSOP is implemented in C++, which compiles natively for the machine, whereas this implementation of \bcubed is implemented in \textit{Java$^{TM}$}. Therefore its performance can suffer from the level of virtualization provided by the \textit{JVM}.
  \item In the standard implementation, SARSOP uses the \textit{Fast Informed Bound} lower heuristic (see section \ref{section:planning:valueFunction}). SARSOP is similar to \bcubed in that as a search algorithm it can be provided with any number of different heuristics to initialize its value function so we modified it to use the $Q_{MDP}$ heuristic so that its results would be more comparable with those of \bcubed. In actuality, the difference between using the two heuristics was not very noticeable for these evaluations.
\end{enumerate}

\begin{table*}[t] 
\centering
    \begin{tabular}{l|c|c|c|c}
             					& \multicolumn{2}{|c|}{\textbf{Rocksample78}}	& \multicolumn{2}{|c}{\textbf{Tag}}	\\ \hline
             \textbf{Algorithm} & \textbf{ADR} & \textbf{Time [ms]} & \textbf{ADR} & \textbf{Time [ms]} 	\\ \hline
             SARSOP 						& 21.28 $\pm$ 0.60 	& 100000* 	& -5.57 $\pm$ 0.52 	& 100000* 	\\
             SARSOP 						& 20.35 $\pm$ 0.58 	& 1000*		& -6.38 $\pm$ 0.52 	& 1000* 	\\
             \bcubed (D=20,$\alpha$=0.95) 	& \textbf{21.60}	$\pm$ 0.02	& 3333 $\pm$ 311 & \textbf{-5.41} $\pm$ 0.06 	& 23798 $\pm$ 716 \\
             \bcubed (D=15,$\alpha$=0.95) 	& 21.49 $\pm$ 0.02  & 1548 $\pm$ 80 & -5.79 $\pm$ 0.06 	& 3878 $\pm$ 124\\
             \bcubed (D=15,$\alpha$=0.65) 	& 21.24 $\pm$ 0.07  & 805 $\pm$ 54	& -5.80 $\pm$ 0.06 	& 1786 $\pm$ 76	\\
             \bcubed (D=10,$\alpha$=0.95) 	& 21.40 $\pm$ 0.06  & 1131 $\pm$ 81	& -6.06 $\pm$ 0.08 	& 521 $\pm$ 33 \\
             \bcubed (D=10,$\alpha$=0.65) 	& 21.14 $\pm$ 0.27  & \textbf{733} $\pm$ 68	& -6.03 $\pm$ 0.10 	& \textbf{317} $\pm$ 24	\\
         \hline
    \end{tabular}
    \caption{Results from \textit{RockSample\_7\_8} and \textit{Tag} for SARSOP and \bcubed in various different configurations. We can see that \bcubed confidently outperforms SARSOP both in reward obtained from domain and convergence time (* means that the algorithm had not converged but was stopped to evaluate policy). The ADR value is provided with a 95\% confidence interval.}
\label{table:planning:rocksample}
\end{table*} 

To evaluate the \bcubed algorithm, we have chosen to use two commonly used POMDP problems called \textit{Rocksample} and \textit{Tag}. We will show anytime performance of \bcubed on these domains, that is how well a policy performs when the algorithm is stopped at an arbitrary time and policy is extracted. We will also compare convergence times with the \textit{Average Discounted Reward} (ADR) measure which is a measure of how much discounted reward one could expect to acquire in a task by acting on a policy produced by the algorithm. All reported times are \textit{on-line} planning times, we do not include the time to calculate the $Q_{MDP}$ heuristic for either system as there exist many different ways to solve MDPs and this does not fall into the domain of the contributions made by either POMDP planners.

\subsection{Rocksample}

Rocksample was introduced by Smith and Simmons to evaluate their algorithm \textit{Heuristic Search Value Iteration} (HSVI) \cite{smith2004heuristic}. In this domain, a robotic rover on mars navigates a grid-world of fixed width and height. Certain (known) grid locations contain rocks which the robot wants to \textit{sample}. Each rock can either have the value \textit{good} or \textit{bad}. If the rover is in a grid location that has a rock, it can \textit{sample} it and receive a reward of 10 if the rock is \textit{good} (in which case sampling it makes the rock turn \textit{bad}) or -10 if the rock was \textit{bad}. The rover can sense any rock in the domain from any grid with a \textit{sense$_i$} action which will return an observation about the rock's value stochastically such that the observation is more accurate the closer the rover is to the rock when it \textit{senses} it. The rover also receives a reward of 10 for entering a terminal grid location on the east side of the map. The rover's location is always fully observable, and the rock locations are static and fully observable but the rock values are initially unknown and only partially observable through the \textit{sense$_i$} action. In \textit{Rocksample\_n\_k}, the world is of size $nxn$ and there are $k$ rocks. The robot can choose from actions \textit{move\_north}, \textit{move\_south}, \textit{move\_east}, \textit{move\_west}, \textit{sample}, \textit{sense$_1$}, \ldots, \textit{sense$_8$}.

\begin{figure}[h!]
	\centering
	\includegraphics[width=3in, trim=1.1in 3in 1.1in 3in] {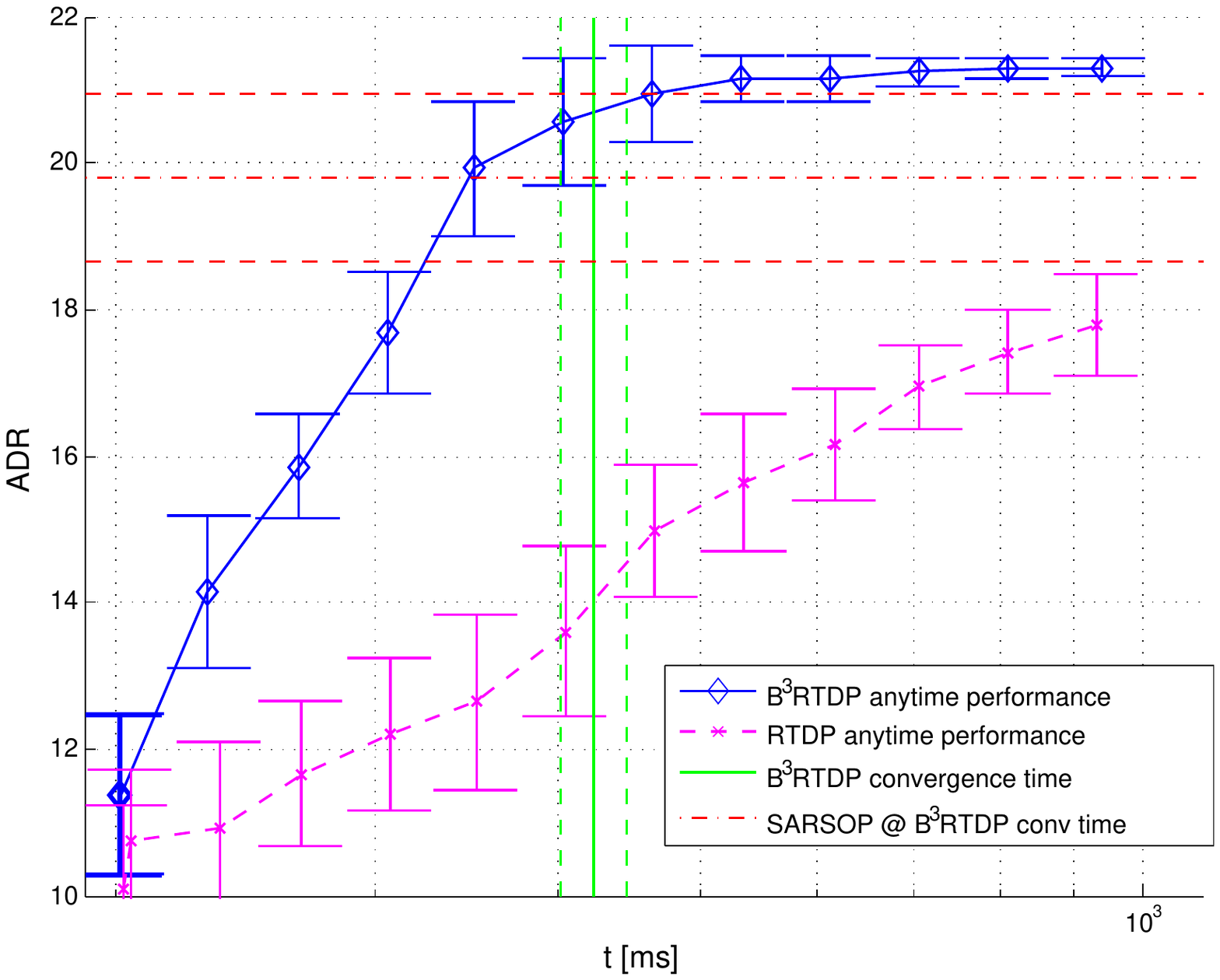}
	\caption{Shows the ADR of \bcubed in the \textit{RockSample\_7\_8} domain. Algorithm was run with $D=10$ and $\alpha=0.75$ and ADR is plotted with error bars showing 95\% confidense intervals calculated from  50 runs.}
	\label{fig:planning:rock_time}
\end{figure}

\subsection{Tag}

\begin{figure}[h!]
	\centering
	\includegraphics[width=3in, trim=1.1in 3in 1.1in 3in] {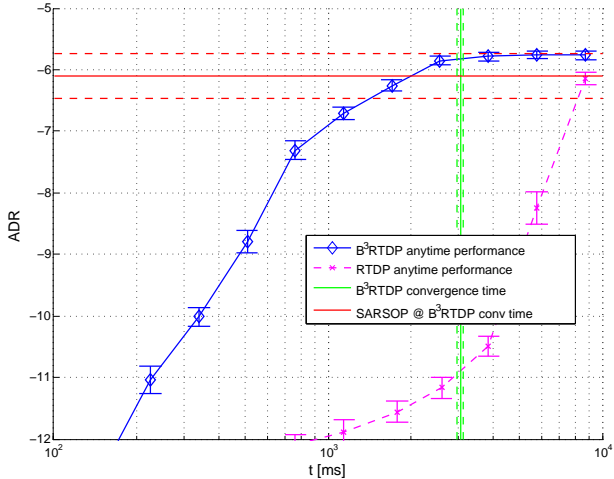}
	\caption{Shows the ADR of \bcubed in the \textit{Tag} domain. The algorithm was run with $D=15$ and $\alpha=0.65$ and ADR is plotted with error bars showing 95\% confidence intervals calculated from  50 runs.}
	\label{fig:planning:tag_anytime}
\end{figure}

The \textit{Tag} domain was introduced by Pineau, Gordon and Thrun to evaluate their \textit{Point-Based Value Iteration} algorithm \cite{pineau2003point}. In this domain, a robot and a human move around a grid-world with a known configuration. The robot's position is fully observable at all times but the human's position can only be observed when the two occupy the same grid. The robot chooses among the following actions: \textit{move\_north}, \textit{move\_south}, \textit{move\_east}, \textit{move\_west} and \textit{tag} and receives a negative reward of -1 for every \textit{move} action, a negative reward of -10 for the \textit{tag} action if it is not in the same grid as the human and positive reward of 10 if it is (which leads to a terminal state). For every \textit{move} action, the human moves away from the robot's position in a stochastic but predictable way.

\begin{figure}[h!]
	\centering
	\includegraphics[width=3in] {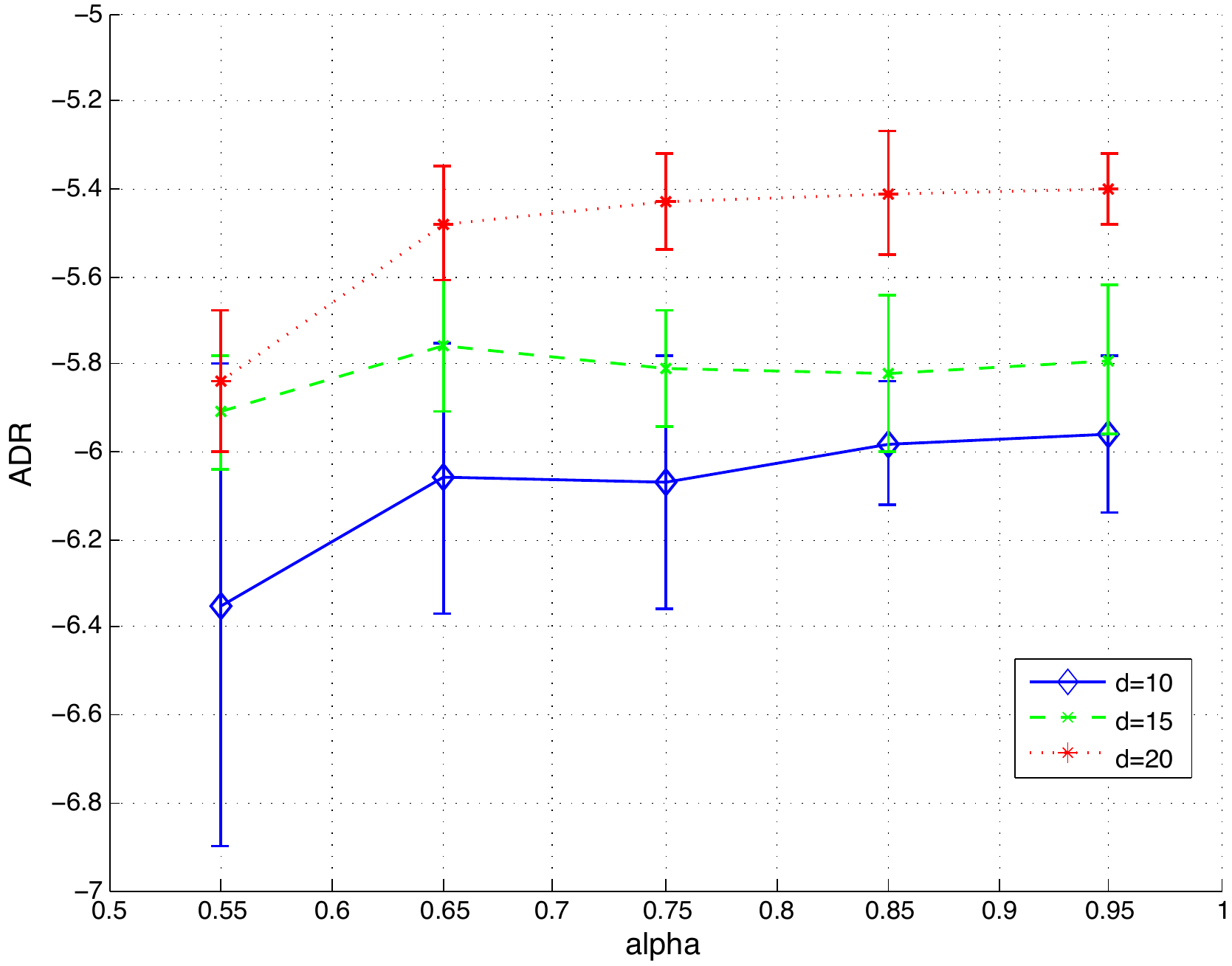}
	\caption{Shows the ADR of \bcubed in the \textit{Tag} domain as a function of the action pruning parameter $\alpha$ and discretization $D$. ADR is plotted with error bars showing a 95\% confidence intervals calculated from  20 runs of the algorithms.}
	\label{fig:planning:tag_parameter_effect1}
\end{figure}

\begin{figure}[h!]
	\centering
	\includegraphics[width=3in] {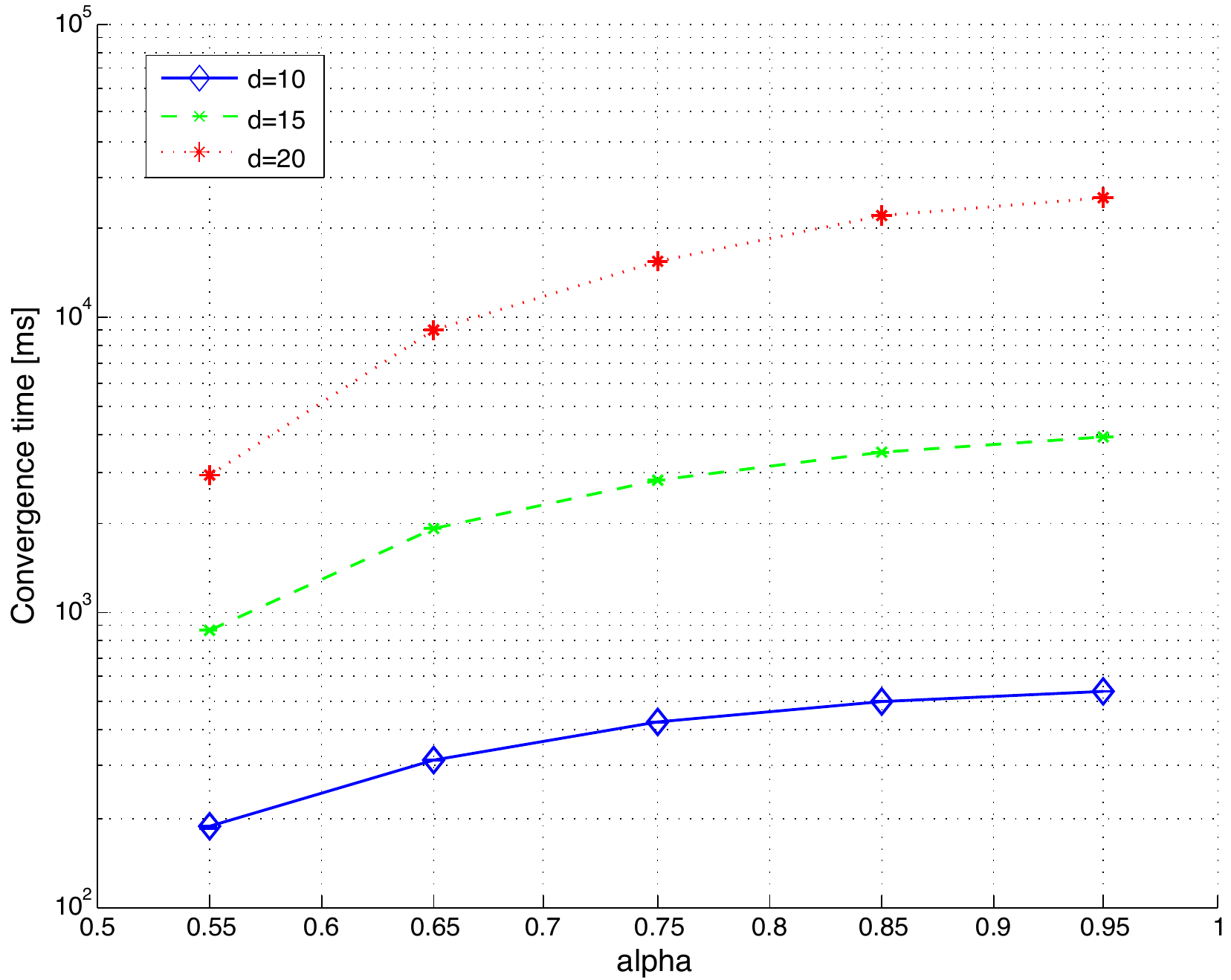}
	\caption{Shows the convergence time of \bcubed in the \textit{Tag} domain as a function of the action pruning parameter $\alpha$ and discretization $D$. ADR is plotted with error bars showing a 95\% confidence interval calculated from  20 runs of the algorithms. We can see that the convergence time of \bcubed increases both with higher discretization as well as a higher requirement of action convergence before pruning. This is an intuitive result as the algorithm also acquires more ADR from the domain in those scenarios.}
	\label{fig:planning:tag_parameter_effect2}
\end{figure}

\section{Discussion and Future Work}\label{section:planning:futureWork}

\subsection{Discussion of Results}
As we can see from table \ref{table:planning:rocksample}, \bcubed can outperform the SARSOP algorithm both in the ADR measure as well as in convergence time. We can also see that the \textit{anytime behavior} of the algorithm is quite good from the graphs in figures \ref{fig:planning:rock_time} and \ref{fig:planning:tag_anytime} such that if the planner were stopped at any time before convergence, it could produce a policy that returns decent reward.

We know that these benefits are largely due to the following factors:
\begin{enumerate}
  \item The belief clustering which is inherent in the discretization scheme of our value function representation. This benefit comes at the cost of memory.
  \item The action pruning that significantly improves convergence time and is enabled by the boundedness of the value function.
  \item The search exploration heuristic which is guided by seeking uncertainty and learning the value function rapidly.
\end{enumerate}

It is satisfying to see such positive results but it should be mentioned that two parameters of the \bcubed algorithm which most heavily impact its performance, namely $D$ and $\alpha$. These parameters have quite domain-dependent implications and should be reconsidered for different domains the algorithm is run on. We show in our results how the performance of the planner varies both in ADR and convergence time as a function of these parameters on the two domains.

\section{Future Work}
During the development of \bcubed we identified several areas where it could be improved with further research and development.

Much of the current running time of the algorithm can be attributed to the belief update calculation of equation \ref{eqn:beliefUpdate}. This update is computationally expensive to carry out or $O(|O||S|^2)$ in the worst case for each action (this can be mitigated by using sparse vector math when beliefs do not have very high entropy). Many POMDP problems have \textit{factored} structure which can be leveraged. This structure means that a state is described by a set of variables, each having their own transition, observation and reward functions. Factored transition functions are traditionally represented as \textit{Dynamic Bayes Nets} (DBNs) and can cause a significant reduction both in the memory requirement of storing the transition matrix as well as in the computational complexity of the belief update. This benefit is gained if the inter-variable dependence of the transition DBNs is not too complex. RTDP-based algorithms would clearly benefit greatly from taking advantage of factored structure, possibly even more so than other algorithms as the \textit{hash table} value function representation might be implemented more efficiently.

To improve convergence of search-based \textit{dynamic programming} algorithms, it is desirable to spend most of the value updates on ``relevant'' and important beliefs. If the point of planning is to learn the true value of the successor beliefs of the initial belief so that a good action choice can be made, then we should prioritize the exploration and value updating of beliefs whose values will make the greatest contribution to that learning. \bcubed already does this to some degree but could do more. The \textit{SARSOP} algorithm \cite{kurniawati2008sarsop} uses a learning strategy where it bins beliefs by discretized features such as the upper value bound and belief entropy. Then, this algorithm uses the average value in the bin as a prediction of the value of a new belief when determining whether to expand it. Focused RTDP \cite{smith2006focused} also attempts to predict how useful certain states are to ``good'' policies and focuses or prioritizes state value updates and tree expansion towards such states. \bcubed could take advantage of the many existing strategies to further focus the belief expansion.

\section{Conclusion}

\textbf{Introduction of a novel general purpose POMDP solver.} We have presented a novel algorithm called \bcubed which extends the \textit{Real Time Dynamic Programming} approach to solving POMDP planning problems. This approach employs a bounded value function representation which it takes advantage of in novel ways. Firstly, it calculates action convergence at every belief and prunes actions that are dominated by others within a certain probability threshold. This technique is similar to a \textit{branch and bound} search strategy but calculates action convergence probabilities such that actions may be pruned before convergence is achieved. Secondly, \bcubed introduces the concept of a \textit{Convergence Frontier} which serves to improve convergence time by taking advantage of convergence of early action selection in the policy. The \bcubed algorithm was evaluated against a state-of-the-art POMDP planner called SARSOP on two standard benchmark domains and showed that it can aquire higher ADR with a shorter convergence time.

\section{Acknowledgments}
This research has been funded by the MIT Media Lab Consortia, and both MURI8 and MURI6 grants from the Office of Naval Research.

% \bibliographystyle{../Authorkit/aaai}
% \bibliography{../library}
% Use this instead for the compiled bibliography.

\end{document}